%% file: root.tex
\documentclass[letterpaper, 10 pt, conference]{ieeeconf}  

\IEEEoverridecommandlockouts                              

\usepackage[utf8]{inputenc} 
\usepackage[T1]{fontenc}    
\usepackage{hyperref}       
\usepackage{url}            
\usepackage{booktabs}       
\usepackage{amsfonts}       
\usepackage{nicefrac}       
\usepackage{microtype}      

\usepackage{booktabs}
\newcommand{\tabincell}[2]{\begin{tabular}{@{}#1 @{}}#2\end{tabular}}
\usepackage{bbm}
\usepackage{amsmath,amsfonts,amssymb,mathtools}
\usepackage{algorithm}
\usepackage{algorithmic}
\usepackage{subfigure}
\usepackage{tikz}
\usepackage{setspace}
\usepackage{multirow}

\usepackage{graphicx}
\usepackage[frozencache=true,cachedir=.]{minted}

\usetikzlibrary{arrows,shapes,chains}

\title{\LARGE \bf
Intrinsically Motivated Self-supervised Learning in Reinforcement Learning}

\author{Yue Zhao$^{1}$, Chenzhuang Du$^{2}$, Hang Zhao$^{2}$, Tiejun Li$^{1, \star}$
\thanks{$^{1}$Peking University, Haidian, Beijing, China}%
\thanks{$^{2}$Tsinghua University, Haidian, Beijing, China}%
\thanks{$^{\star}$Correspondence author {\tt\small tieli@math.pku.edu.cn}}%
}

\newcommand{\rebuttal}[1]{{{#1}}}

\begin{document}

\maketitle
\thispagestyle{empty}
\pagestyle{empty}

\input{sec0_abs}
\input{sec1_intro}

\input{sec2_method}

\input{sec3_analysis}
\input{sec4_exp}

\input{sec5_related}

\input{sec6_conclusion}


\input{appendix}

\addtolength{\textheight}{-12cm}   

\bibliographystyle{IEEEtran}
\bibliography{ref}  


\end{document}

%% file: sec0_abs.tex
\begin{abstract}
In vision-based reinforcement learning (RL) tasks, it is prevalent to assign auxiliary tasks with a surrogate self-supervised loss so as to obtain more semantic representations and improve sample efficiency. However, abundant information in self-supervised auxiliary tasks has been disregarded, since the representation learning part and the decision-making part are separated.
To sufficiently utilize information in auxiliary tasks, we present a simple yet effective idea to employ self-supervised loss as an intrinsic reward, called \textbf{I}ntrinsically \textbf{M}otivated \textbf{S}elf-\textbf{S}upervised learning in \textbf{R}einforcement learning (IM-SSR). 
We formally show that the self-supervised loss can be decomposed as {exploration} for novel states and {robustness} improvement from nuisance elimination.
IM-SSR can be effortlessly plugged into any reinforcement learning with self-supervised auxiliary objectives with nearly no additional cost.
Combined with IM-SSR, the previous underlying algorithms achieve salient improvements on both sample efficiency and generalization in various vision-based robotics tasks from the DeepMind Control Suite, especially when the reward signal is sparse.

\end{abstract}

%% file: sec1_intro.tex
\section{Introduction}

Reinforcement learning has achieved significant success in many fields with visual observations~\cite{hansen2020generalization, srinivas2020curl, mnih2015human,schmeckpeper2020reinforcement}.
Sample efficiency and representation learning in vision-based reinforcement learning tasks have, though, hitherto been a challenging problem~\cite{srinivas2020curl, yarats2019improving}, especially when applying RL into real-world robotics~\cite{ibarz2021train}. Assigning an auxiliary task with a surrogate loss, which introduces the powerful and promising self-supervised learning (SSL) methods in computer vision to RL, is an efficacious technique for representation learning and sample efficiency. With self-supervised auxiliary objectives, the RL algorithms learn more semantic representations and \rebuttal{achieve} faster convergence. 

Albeit combining with SSL naively can improve sample efficiency in vision-based RL, there is still room for improvements. Especially in a real-world setting with sparse rewards, exploration becomes desiderata for RL to obtain non-trivial reward signals, which remains a predicament that previous SSL-RL baselines can not solve efficiently. Adequate information in self-supervised auxiliary tasks can offer practical assistance; however, such paramount information has been disregarded, since the representation learning part is separated from the decision-making part.

Based on that,
we propose a framework for SSL-RL to assign the surrogate self-supervised loss into policy learning as an intrinsic reward, namely \textbf{I}ntrinsically \textbf{M}otivated \textbf{S}elf-\textbf{S}upervised learning in \textbf{R}einforcement learning (IM-SSR). It is a simple yet effective modification with nearly no additional cost, which can be effortlessly plugged in any reinforcement learning with self-supervised auxiliary tasks. 
Intuitively, the self-supervised loss measures the quality of the observation representation learning, which would be relatively large for the states rarely encountered or vulnerable to nuisance. Also we theoretically decompose the self-supervised loss as a metric in the feature space: one motivates exploration for novel states, the other improves robustness from nuisance elimination. Thereby, it is reasonable to design the self-supervised loss as an intrinsic reward to award extra bonuses for those states. Combining with this intrinsic reward, the agent is encouraged to explore novel and vulnerable states, \rebuttal{thus improving} sample efficiency along side generalization ability.

Empirical experiments are conducted on robotics control tasks in DeepMind Control Suite~\cite{tassa2018deepmind} based on a generalization benchmark DMControl-GB\footnote{https://github.com/nicklashansen/dmcontrol-generalization-benchmark}. Complex testing tasks are included, like environments with unseen natural video backgrounds, which \rebuttal{can evaluate the generalization ability of the algorithm} to various real-world settings.
Two of the classical vision-based reinforcement learning algorithms with self-supervised methods, CURL \cite{srinivas2020curl} and SODA \cite{hansen2020generalization}, are chosen to be baselines. Combined with IM-SSR, \rebuttal{the} underlying baselines improve sample efficiency and generalization, especially when the reward signals are sparse, \rebuttal{resulting in} new state-of-the-art performance.

We summarize our contributions as follows:
\begin{itemize}
    \item We present a general framework, IM-SSR, which is an effective idea to utilize the self-supervised loss in auxiliary tasks as an intrinsic reward. The philosophy of associating self-supervised learning with reinforcement learning is promising since the information in auxiliary tasks should not be disregarded.
    \item We formally show that the self-supervised loss can be decomposed as \textbf{exploration}  for novel states and \textbf{robustness} in representations from nuisance elimination.
    \item With nearly no additional cost, we offer an approach to promote improvements on the previous underlying reinforcement learning algorithms with self-supervised auxiliary loss, which achieves new state-of-the-art performance on vision-based reinforcement learning. 
    
\end{itemize}

%% file: sec2_method.tex
\section{Methodology}

\begin{figure*}[t]
  \centering
  \includegraphics[width=13.5cm]{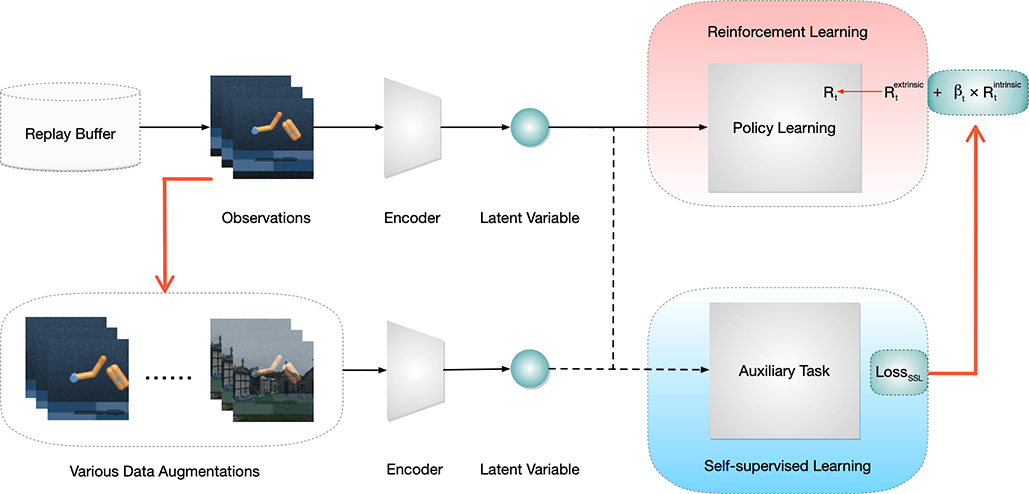}
  \caption{Intrinsically Motivated Self-supervised Learning in Reinforcement Learning Framework}
  \label{sslrl}
\end{figure*}
As demonstrated in Figure \ref{sslrl}, the framework of reinforcement learning combined with self-supervised learning is: First, we maintain a replay buffer with transitions. 
Various kinds of augmentations can be employed to obtain different views of the original observations. It remains ambiguous whether to use the original observation or \rebuttal{an augmented view} into the encoder for RL, which we demonstrate as above just for simplicity. In reinforcement learning part, we encode inputs into latent variable $z$ and train the policy based on $z$ and related transitions; in self-supervised learning part, an auxiliary task is conducted to \rebuttal{constrain} the consistency of $z^\prime$ encoded from different views of observations. The self-supervised loss in auxiliary task is generally denoted as $\mathcal{L}_{\text{SSL}}$ \rebuttal{and explicitly formulated in Appendix~\ref{ssl-formulation}}. To utilize the self-supervised learning part, we introduce $\mathcal{L}_{\text{SSL}}^{i}$ on each observation as an intrinsic reward $ R^{intrinsic}$ added to the original extrinsic reward in Equation (\ref{reward}).
\begin{equation}
    \label{reward}
    R_t = R^{extrinsic}_t + \beta_t * R^{intrinsic}_t
\end{equation}
 When $\beta_t = 0$, i.e., $R_t = R_t^{extrinsic}$, the framework degenerates into previous SSL-RL methods, which have no interplay between SSL and RL.

\textbf{Motivation.} Intuitively, the self-supervised loss contains information about \rebuttal{how well the learned representation is}. The loss would be relatively large for rarely encountered states, which motivates us to award novel or vulnerable states. 
The bonus motivates the agent to explore, 
which helps with the searching for non-trivial rewards. Based on that, sample efficiency can be improved significantly, especially in sparse-rewarded tasks. We have to clarify that IM-SSR utilizes the inherent yet unemployed information in SSL, which is different from those manually designed exploration methods. Also, generalization ability is a by-product gained from building robust representations for vulnerable states.

\textbf{Summary.} 
To utilize inherent information in SSL, we present a generic framework, IM-SSR, where the SSR can be replaced by any suitable self-supervised learning method. 
For instance, IM-CURL means a CURL baseline combined with the intrinsic reward modification. 
 Our method requires no additional network architectures nor any modification on the self-supervised task, thus making it easy to be plugged-in.

The pseudo-code of IM-SSR is as follows.

\begin{algorithm}[H]
  \caption{IM-SSR Framework}
  \label{alg:example}
\begin{algorithmic}
  \STATE {\bfseries Input:} Replay buffer $\mathcal{B}$, RL update frequency $m$, RL batch size $N$, Intrinsic parameter $\beta$
  \REPEAT
  \FOR{$0$ {\bfseries to} $m$}
  \STATE Sample a batch of $N$ transitions $\{(o_i, a_i, o_i^\prime, r_i, \text{done}_i)\}^N_{i=1} \sim \mathcal{B}$
  \STATE Compute self-supervised loss $l_i$ between each observation and its augmentation
  \STATE Normalize within the batch: $\{r^{intrinsic}_i\}^N_{i=1} \leftarrow \texttt{Normalize}(\{l_i\}^N_{i=1})$
  \STATE Add intrinsic reward: $r_i \leftarrow r_i + \beta * r^{intrinsic}_i$
  \STATE Optimize reinforcement learning objective $\mathcal{L}_\text{{RL}}$
  \ENDFOR
  \STATE Sample a new batch of transitions from $\mathcal{B}$ or use transitions sampled in RL updates
  \STATE Optimize self-supervised learning objective $\mathcal{L}_\text{{SSL}}$
  \UNTIL{reach max training steps}
\end{algorithmic}
\end{algorithm}

\textbf{Implementation details.} The intrinsic rewards need to be normalized, since it is the relative value inside one batch/update that matters, not the exact quantity. The hyper-parameter $\beta_t$ can be elaborately designed to decay; however, we implement the method with a fixed $\beta$ or a naive decaying schedule without fine-tuning and it still achieves desired performance. Our method can be easily deployed on any SSL-RL framework.
In the experiments, we strictly follow the same way as the corresponding baseline does and use exactly the same augmentation method. Details of the related SSL-RL baselines are shown in Appendix~\ref{ssl-formulation}.

%% file: sec3_analysis.tex
\section{Analysis on Self-supervised Loss in Reinforcement Learning}\label{analysis}

We formally show that it is reasonable to employ the self-supervised loss as an intrinsic reward, which can be interpreted as an exploration bonus and robustness improvement from nuisance elimination. 

\subsection{Notations and Preliminaries}

\begin{figure}[t]
  \centering
	\tikzstyle{format}=[rectangle,thin,fill=white]
	\tikzstyle{agent}=[rectangle,aspect=2,draw,thin] 
	\tikzstyle{test}=[diamond,aspect=10,draw,thin]
	\tikzstyle{point}=[coordinate,on grid,]
	\begin{tikzpicture}[node distance=8mm,
			auto,>=latex',
			thin,
			start chain=going right,
			every join/.style={norm},]
		\node[format] (n0){$n_t$};
		\node[format,below of=n0] (n1){$x_t$};
		\node[format,right of=n1, node distance=10mm] (n2){$o_t$};
		\node[format,right of=n2, node distance=10mm] (n3){$z_t$};
		\node[agent,right of=n3,node distance=14mm] (n4){Agent};
		\node[format, right of=n0,node distance=52mm] (n5){$n_{t+1}$};
		\node[format,below of=n5, node distance=8mm] (n6){$x_{t+1}$};
		\node[format,right of=n6, node distance=13mm] (n7){$o_{t+1}$};
		\node[format,right of=n7, node distance=13mm] (n8){$z_{t+1}$};
		\draw[->] (n0.south) -- (n2);
		\draw[->] (n1.east) -- (n2);
		\draw[->] (n2.east) -- (n3);
		\draw[->] (n3.east) -- (n4);
		\draw[->,dashed] (n4.east) to node{$a_t$\ } (n6);
		\draw[->] (n5.south) -- (n7);
		\draw[->] (n6.east) -- (n7);
		\draw[->] (n7.east) -- (n8);
	\end{tikzpicture}
    \caption{Representation MDP.}
  \label{rep_mdp}
\end{figure}
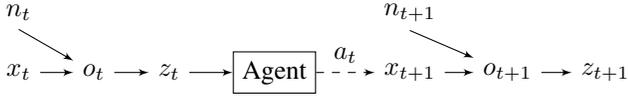

A Markov Decision Process~(MDP) is defined for the information transmission in Figure \ref{rep_mdp}. $x \in \mathcal{X}$ denotes an optimal representation in the latent space, which is necessary and sufficient for the reinforcement learning task $y$. Another random variable $n \in \mathcal{N}$ denotes nuisance for the task, i.e., $n \perp y$ or $I(y;n) = 0$; also, we assume $n \perp x$ and $I(x;n) = 0$. Observation $o$ is generated by an implicit function $g(\cdot, \cdot)$, i.e., $o = g(x, n) \in \mathcal{O}$. For instance, $o$ denotes the stacked pixel observation captured from the environment; $x$ is the desired optimal representation for downstream tasks, such as state-based proprioceptive features; $n$ denotes task-irrelevant information like textures, shadows and backgrounds, which may be a culprit for the representation learning, thus seriously depreciating sample efficiency and generalization ability to unseen environments of the algorithm.

We now consider a generic setting for contrastive learning as an auxiliary self-supervised task along side reinforcement learning. 
The encoding function is denoted by $f: \mathcal{O} \rightarrow \mathcal{Z}$. A main principle for contrastive learning is that features from different views of the same observation should be forced to be close in the feature space, i.e., the ideal encoder $f^\star$ would satisfy{ $\text{dist}(f^\star(o_i), f^\star(o_j)) \propto \text{dist}(x_i, x_j)$}. The contrastive loss depends on the SSL architecture, which {has the form like} Equation~(\ref{infonce}) in CURL or Equation~(\ref{sodaloss}) in SODA {as shown in Appendix~\ref{ssl-formulation}}. Then we use ${\rho}(z_i, z_j)$ to denote the underlying pair-wise contrastive loss as a metric defined on a metric space $\mathcal{H}$, measuring the distance between different latent representations $z_i \in \mathcal{Z}$. It satisfies the well-known properties: $\rho(x, y) \geq 0$; $\rho(x, y) = \rho(y, x)$; $\rho(x, y) \leq \rho(x, z) + \rho(y, z)$.

\subsection{Decomposition and Interpretation of Contrastive Loss}
For each observation $o = g(x, n)$, we analyze the pair-wise contrastive loss. Augmentations of $o$ are denoted by $o^\prime = t(o)$. Since the augmentation function is designed to perturb nuisance and maintain significant information of the original observation, we assume $\exists\ n^\prime$, s.t. $o^\prime = t(o) = g(x, n^\prime)$. 
The corresponding features are $z = f(o)$ and $z^\prime = f(o^\prime)$. We further introduce an optimal representation function $f^\star$. Ideally, it would ignore information related to nuisance $n$ and maintain the information in $x$ related to downstream tasks, i.e., $f^\star(g(x, n)) = f^\star(g(x, n')), \forall n, n' \in \mathcal{N}$. We may use a fixed $n_0$ to denote $z^\star = f^\star(g(x, n)) = f^\star(g(x, n_0)), \forall n\in \mathcal{N}$.
\begin{equation}
\begin{aligned}
&\rho({z}^\prime , z) = \rho(f(g(x, n^\prime)), f(g(x, n))) \\
\leq &\rho(f(g(x, n^\prime)), f^\star(g(x, n_0))) + \rho(f(g(x, n)), f^\star(g(x, n_0)))  \\
= &\rho(h(x, n^\prime), h^\star(x, n_0)) + \rho(h(x, n), h^\star(x, n_0)) ,
\end{aligned}
\end{equation}
The loss can be decomposed as above, where $h / h^\star$ denotes the composite function of $g$ and $f / f^\star$, and the second inequality is from the triangle inequality of the metric. Each term on the right hand side can be further bounded by

\begin{equation}
\begin{aligned}
& \rho(h(x, n), h^\star(x, n_0)) \\
\leq & \underbrace{\rho(h(x, n_0), h^\star(x, n_0))}_{\text{(i) projected distance of x}} + \underbrace{\rho(h(x, n), h(x, n_0))}_{\text{(ii) nuisance elimination}} .
\end{aligned}
\end{equation}

The first term, interpreted as the projected distance of $x$, is indeed an exploration bonus. It is the distance in feature space between different projection functions $h(., n_0)$ and $h^\star(., n_0)$, which contrastive learning is trying to minimize; therefore, the distance would be smaller if the state has been encountered. From the perspective of exploration in reinforcement learning, this term can be treated as a prediction error, and it would bring extra bonus for novel states that the contrastive part is not trained on. The prediction error is often considered to motivate exploration like RND~\cite{burda2018exploration} does; however, it requires additional networks for distillation. In the most desired case, this term should vanish, as the function $h$ is optimized to $h^\star$. 

The second term, related to nuisance elimination, encourages the agent to visit vulnerable states and introduces randomness into the optimization. States that are more sensitive to distortions will give larger distance between $h(x, n)$ and $h^\star(x, n_0)$, which is encouraged to be visited or revisited. To be clarified, since the nuisance $n$ in the considered MDP is independent of $x$, our method can not directly improve the generalization ability of the baseline; however, it can contribute by building more robust representations for vulnerable states. 
Robustness in representations makes the learned representations more invariant to distortions, thus improving the generalization ability to untrained environments.
While the function $h$ converges to $h^\star$, the information in $n$ is getting ignored, and the whole term of nuisance elimination would also vanish. 

To summarize, the intrinsic reward, i.e., pair-wise contrastive loss of the auxiliary tasks in reinforcement learning, can be interpreted as: \textbf{exploration} for novel states that improves sample efficiency, and \textbf{robustness} in representations from nuisance elimination that improves generalization ability.
\begin{equation}
\begin{aligned}
\label{finalbound}
 \rho({z}^\prime , z)  &
\leq \underbrace{2\rho(h(x, n_0), h^\star(x, n_0))}_{\text{(i) \textbf{Exploration} for novel states}} \\ + &\underbrace{\left[\rho(h(x, n), h(x, n_0)) + \rho(h(x, n^\prime), h(x, n_0))\right]}_{\text{(ii) \textbf{Robustness} from nuisance elimination}}.
\end{aligned}
\end{equation}
Inequality~(\ref{finalbound}) is tightly bounded, since $h$ is optimized towards $h^\star$ and $h^\star$ will ignore the information in $n$ ideally. Apart from that, with a decaying parameter $\beta_t$ in front of the intrinsic reward, our method is able to converge to the optimal policy asymptotically. Specifically, $\rho({z}^\prime , z) $ can be a pair-wise loss of Equation~(\ref{infonce}) in CURL or Equation~(\ref{sodaloss}) in SODA, which will serve as an intrinsic reward in IM-SSR. The pair-wise loss can also be designed by ourselves to utilize the information in encoders, which will be further explored in Appendix~\ref{app:imsvea}.

%% file: sec4_exp.tex
\section{Experiments}

\renewcommand\arraystretch{0.85}
\begin{table*}[htbp]
\caption{The results of CURL, SODA, IM-CURL and IM-SODA in unmodified environments.}
 \label{sample}
  \resizebox{\textwidth}{!}
  {
    \begin{tabular}{lcc|cc||cc|cc||cc|cc}
    \toprule
    \multicolumn{1}{c}{\multirow{2}[2]{*}{\tabincell{c}{DMControl Suite \\ \texttt{(training)}}}} & \multicolumn{4}{c||}{\textbf{0.2T} Training Frames}     & \multicolumn{4}{c||}{\textbf{0.5T} Training Frames}    & \multicolumn{4}{c}{\textbf{1.0T} Training Frames} \\
    
    \cmidrule{2-13} 
    & \multicolumn{1}{c}{CURL} & \multicolumn{1}{c|}{IM-CURL} & \multicolumn{1}{c}{SODA} & \multicolumn{1}{c||}{IM-SODA} & \multicolumn{1}{c}{CURL} & \multicolumn{1}{c|}{IM-CURL} & \multicolumn{1}{c}{SODA} & \multicolumn{1}{c||}{IM-SODA} &\multicolumn{1}{c}{CURL} & \multicolumn{1}{c|}{IM-CURL} & \multicolumn{1}{c}{SODA} & \multicolumn{1}{c}{IM-SODA}\\

   \midrule
    \tabincell{l}{{\texttt{cartpole}} \\ {\texttt{swingup\_sparse}}} & \tabincell{c}{50.1 \\ \footnotesize{$\pm$50.3} } & \tabincell{c}{\textbf{318.4} \\ \footnotesize{$\pm$240.3}} & \tabincell{c}{6.6 \\ \footnotesize{$\pm$11.1}} & \tabincell{c}{\textbf{10.3} \\ \footnotesize{$\pm$11.7}} & \tabincell{c}{ \textbf{726.1}\\ \footnotesize{$\pm$48.0}} & \tabincell{c}{721.9 \\ \footnotesize{$\pm$48.7}} & \tabincell{c}{435.5 \\ \footnotesize{$\pm$361.9}} & \tabincell{c}{\textbf{632.1} \\ \footnotesize{$\pm$249.2}} & \tabincell{c}{ 744.2\\ \footnotesize{$\pm$23.8}} & \tabincell{c}{\textbf{744.8} \\ \footnotesize{$\pm$54.2}} & \tabincell{c}{617.3 \\ \footnotesize{$\pm$331.4}} & \tabincell{c}{\textbf{792.0} \\ \footnotesize{$\pm$38.5}}\\

   \specialrule{0em}{2pt}{2pt} 
    
   \tabincell{l}{\texttt{finger} \\ \texttt{spin}} & \tabincell{c}{784.2 \\ \footnotesize{$\pm$112.4} } & \tabincell{c}{ \textbf{789.9}\\ \footnotesize{$\pm$94.2}} & \tabincell{c}{779.5 \\ \footnotesize{$\pm$108.8}} & \tabincell{c}{\textbf{861.8} \\ \footnotesize{$\pm$76.8}} & \tabincell{c}{862.3 \\ \footnotesize{$\pm$132.7}} & \tabincell{c}{ \textbf{916.8}\\ \footnotesize{$\pm$37.4}} & \tabincell{c}{909.3 \\ \footnotesize{$\pm$50.8}} & \tabincell{c}{\textbf{975.9} \\ \footnotesize{$\pm$8.2}} & \tabincell{c}{ 920.9\\ \footnotesize{$\pm$124.8}} & \tabincell{c}{ \textbf{949.7}\\ \footnotesize{$\pm$39.4}}  & \tabincell{c}{937.1 \\ \footnotesize{$\pm$35.3}} & \tabincell{c}{\textbf{983.1} \\ \footnotesize{$\pm$2.4}} \\
   
    \specialrule{0em}{2pt}{2pt} 
    
    \tabincell{l}{{\texttt{pendulum}} \\ {\texttt{swingup}}} & \tabincell{c}{\textbf{22.6} \\ \footnotesize{$\pm$38.3} } & \tabincell{c}{21.5 \\ \footnotesize{$\pm$41.4}} & \tabincell{c}{22.3 \\ \footnotesize{$\pm$34.1}} & \tabincell{c}{\textbf{36.5} \\ \footnotesize{$\pm$38.5}} & \tabincell{c}{ 41.8\\ \footnotesize{$\pm$46.0}} & \tabincell{c}{ \textbf{239.5}\\ \footnotesize{$\pm$285.3}} & \tabincell{c}{175.4 \\ \footnotesize{$\pm$222.2}} & \tabincell{c}{\textbf{494.3} \\ \footnotesize{$\pm$359.5}} & \tabincell{c}{ 171.9\\ \footnotesize{$\pm$}289.8} & \tabincell{c}{\textbf{780.3} \\ \footnotesize{$\pm$38.9}} & \tabincell{c}{474.1 \\ \footnotesize{$\pm$391.0}} & \tabincell{c}{\textbf{{736.2}} \\ \footnotesize{$\pm$196.1}}\\

   \specialrule{0em}{2pt}{2pt}  
   
   \tabincell{l}{\texttt{reacher} \\ \texttt{easy}} & \tabincell{c}{{871.1} \\ \footnotesize{$\pm$97.0} } & \tabincell{c}{\textbf{877.6} \\ \footnotesize{$\pm$50.1}} & \tabincell{c}{474.9 \\ \footnotesize{$\pm$56.5}} & \tabincell{c}{\textbf{513.5} \\ \footnotesize{$\pm$127.8}} & \tabincell{c}{\textbf{952.7} \\ \footnotesize{$\pm$36.8}} & \tabincell{c}{904.7 \\ \footnotesize{$\pm$86.7}} & \tabincell{c}{472.0 \\ \footnotesize{$\pm$179.2}} & \tabincell{c}{\textbf{644.7} \\ \footnotesize{$\pm$153.2}} & \tabincell{c}{932.6 \\ \footnotesize{$\pm$46.5}} & \tabincell{c}{\textbf{970.0} \\ \footnotesize{$\pm$4.9}}  & \tabincell{c}{539.8 \\ \footnotesize{$\pm$285.2}} & \tabincell{c}{\textbf{709.7} \\ \footnotesize{$\pm$169.7}} \\

   \specialrule{0em}{2pt}{2pt} 
    
   \tabincell{l}{\texttt{walker} \\ \texttt{walk}} & \tabincell{c}{{680.3} \\ \footnotesize{$\pm$79.5} } & \tabincell{c}{\textbf{766.7}\\ \footnotesize{$\pm$107.7}} & \tabincell{c}{618.1 \\ \footnotesize{$\pm$65.6}} & \tabincell{c}{\textbf{641.7} \\ \footnotesize{$\pm$43.3}} & \tabincell{c}{813.4 \\ \footnotesize{$\pm$63.4}} & \tabincell{c}{\textbf{832.4} \\ \footnotesize{$\pm$118.1}} & \tabincell{c}{735.2 \\ \footnotesize{$\pm$61.3}} & \tabincell{c}{\textbf{776.3} \\ \footnotesize{$\pm$41.5}} & \tabincell{c}{855.3 \\ \footnotesize{$\pm$66.8}} & \tabincell{c}{\textbf{873.8} \\ \footnotesize{$\pm$88.0}} & \tabincell{c}{\textbf{840.3} \\ \footnotesize{$\pm$47.2}} & \tabincell{c}{{834.9} \\ \footnotesize{$\pm$52.1}}\\
   
    \bottomrule
    \end{tabular}%
  }
\end{table*}%

\begin{figure*}[t]

  \centering
  \subfigure[\texttt{training}]{\label{training}\includegraphics[scale=0.4]{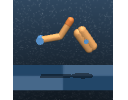}} \hspace{.15in}
  \subfigure[\texttt{color\_easy}]{\includegraphics[scale=0.4]{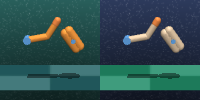}}\hspace{.15in}
  \subfigure[\texttt{color\_hard}]{\label{color-hard}\includegraphics[scale=0.4]{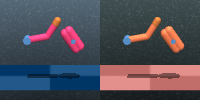}}\hspace{.15in}
  \subfigure[\texttt{video\_easy}]{\includegraphics[scale=0.4]{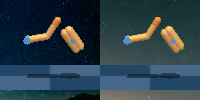}}\hspace{.15in}
    \subfigure[\texttt{video\_hard}]{\label{video-hard}\includegraphics[scale=0.4]{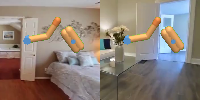}}
  \caption{DMControl-GB Environments \cite{hansen2020generalization}}
  \label{fig:envs}
\end{figure*}

\subsection{Experimental Settings}
Empirically, we conduct experiments on DMControl suite. 
All experiments are evaluated over 5 random seeds, except for those mentioned in the related caption.
DMControl-GB provides diverse test environments for policy evaluation, which are important for {sim-to-real} in robotics. The agent is trained in the unmodified \texttt{training} environment, while it is evaluated in some new environments as well as the unmodified environment of DMControl suite. The new test environments can be summarized as two types: changing the color randomly and replacing the background with natural videos. Specifically, it includes \texttt{color\_easy}, \texttt{color\_hard}, \texttt{video\_easy}, \texttt{video\_hard}, as shown in Figure~\ref{fig:envs} and we select \texttt{training}, \texttt{color\_hard} and \texttt{video\_hard} as the representative environments.

Primarily, to validate the priority in sample efficiency of IM-SSR, we compare IM-CURL, IM-SODA with CURL and SODA in the \texttt{training} environment. 
Especially, we elaborately design tasks with sparse rewards to further prove the significant exploration brought by IM-SSR. We conduct all experiments with 500K training frames. We denote the total training frames as \textbf{T}, using {0.2T} as early-term, {0.5T} as mid-term and {1.0T} as final-term to represent the training progress of 20\%, 50\% and 100\%. 
We report the mean and standard deviation in the tables.
Additional results and more details can be found in Appendix~\ref{ref-performance} and \ref{ref-implementation}.

Besides, to verify that IM-SSR is helpful for robustness and generalization, we compare IM-SODA and SODA in challenging generalization environments. First, in Section~\ref{section-gen}, we intend to evaluate the generalization ability of sim-to-real, where agents are trained in simulated unmodified mujoco environment \texttt{training}, and tested in real-world like environment \texttt{color\_hard} and \texttt{video\_hard}. Second, we design tasks which are trained on one scene and tested on another to further validate the generalization in transferring. The detailed analysis of generalization tasks and results can be found in Appendix~\ref{transfer-description}.

\subsection{Sample Efficiency}

In this subsection, we demonstrate that IM-SSR can improve sample efficiency on the \texttt{training} environment as Figure~\ref{training}. The results shown in Table~\ref{sample} can be summarized: 
\begin{itemize}
    \item IM-SSR surpasses its underlying baselines in the early-term in most cases and maintains its priority in the mid-term significantly. 
    \item The improvements in mid-term are much larger than in other terms, since IM-SSR converges much faster than underlying baselines. The baseline may still struggle in learning representations or searching for non-trivial rewards, while IM-SSR has already learned semantic representations and obtained positive and effective reward signals to be trained on.
    \item In the final-term, we expect IM-SSR and baselines to achieve similar results like in \texttt{walker\_walk}. However, we surprisingly find that in several tasks like \texttt{cartpole\_swingup\_sparse}, IM-SSR achieves higher mean and much lower variance in performance than the underlying baselines, which means IM-SODA is much more stable in these environments.
    \item There are few cases that IM-SSR provides slight improvements, which mainly attribute to the dense reward. When setting the reward signal to be sparse, the improvements of IM-SSR become significant. Related experiments are implemented in Section~\ref{section:sparse}.
\end{itemize}

\subsection{Generalization}\label{section-gen}

We demonstrate that IM-SSR can improve robustness and generalization, and 
the evaluation is conducted on \texttt{color\_hard} and \texttt{video\_hard} environments as Figure \ref{color-hard} and Figure \ref{video-hard}.

\renewcommand\arraystretch{0.85}
\begin{table*}[htpb]
\caption{Generalization results of SODA and IM-SODA on \texttt{color\_hard} and \texttt{video\_hard}.}
  \label{generalization}
  \resizebox{\textwidth}{!}
  {
    \begin{tabular}{lcc|cc|cc||cc|cc|cc}
        \multicolumn{1}{c}{\textbf{ }} & \multicolumn{6}{c}{\textbf{random colors}} & \multicolumn{6}{c}{\textbf{video backgrounds}} \\
    \toprule
    \multicolumn{1}{c}{\multirow{2}[2]{*}{\tabincell{c}{DMControl-GB \\ \texttt{(evaluating)}}}} & \multicolumn{2}{c|}{\textbf{0.2T} }     & \multicolumn{2}{c|}{\textbf{0.5T} }    & \multicolumn{2}{c||}{\textbf{1.0T} }  & \multicolumn{2}{c|}{\textbf{0.2T} }     & \multicolumn{2}{c|}{\textbf{0.5T} }    & \multicolumn{2}{c}{\textbf{1.0T} } \\
    
    \cmidrule{2-13} 
    & \multicolumn{1}{c}{SODA} & \multicolumn{1}{c|}{IM-SODA} & \multicolumn{1}{c}{SODA} & \multicolumn{1}{c|}{IM-SODA} & \multicolumn{1}{c}{SODA} & \multicolumn{1}{c||}{IM-SODA} & \multicolumn{1}{c}{SODA} & \multicolumn{1}{c|}{IM-SODA} &\multicolumn{1}{c}{SODA} & \multicolumn{1}{c|}{IM-SODA} & \multicolumn{1}{c}{SODA} & \multicolumn{1}{c}{IM-SODA}\\

   \midrule
    \tabincell{l}{{\texttt{cartpole}} \\ {\texttt{swingup\_sparse}}} & \tabincell{c}{5.8 \\ \footnotesize{$\pm$9.1} } & \tabincell{c}{\textbf{10.7} \\ \footnotesize{$\pm$11.3}} & \tabincell{c}{341.9 \\ \footnotesize{$\pm$282.4}} & \tabincell{c}{\textbf{429.0} \\ \footnotesize{$\pm$223.5}} & \tabincell{c}{468.6 \\ \footnotesize{$\pm$257.1}} & \tabincell{c}{ \textbf{573.8}\\ \footnotesize{$\pm$104.2}} & \tabincell{c}{1.5 \\ \footnotesize{$\pm$2.5}} & \tabincell{c}{\textbf{4.9} \\ \footnotesize{$\pm$5.3}} & \tabincell{c}{73.5\\ \footnotesize{$\pm$}55.7} & \tabincell{c}{\textbf{89.8}\\ \footnotesize{$\pm$}69.7} & \tabincell{c}{111.9 \\ \footnotesize{$\pm$77.5}} & \tabincell{c}{\textbf{139.8} \\ \footnotesize{$\pm$86.7}}\\

   \specialrule{0em}{2pt}{2pt} 
    
   \tabincell{l}{\texttt{finger} \\ \texttt{spin}} & \tabincell{c}{717.7 \\ \footnotesize{$\pm$109.3} } & \tabincell{c}{\textbf{811.5} \\ \footnotesize{$\pm$74.7}} & \tabincell{c}{831.5 \\ \footnotesize{$\pm$48.7}} & \tabincell{c}{\textbf{925.1} \\ \footnotesize{$\pm$50.1}} & \tabincell{c}{879.2 \\ \footnotesize{$\pm$40.8}} & \tabincell{c}{\textbf{931.2} \\ \footnotesize{$\pm$37.4}} & \tabincell{c}{354.6 \\ \footnotesize{$\pm$50.9}} & \tabincell{c}{\textbf{361.1} \\ \footnotesize{$\pm$64.2}} & \tabincell{c}{417.8 \\ \footnotesize{$\pm$57.3}} & \tabincell{c}{\textbf{481.4} \\ \footnotesize{$\pm$31.6}}  & \tabincell{c}{333.6 \\ \footnotesize{$\pm$18.1}} & \tabincell{c}{\textbf{353.6} \\ \footnotesize{$\pm$11.4}} \\
   
    \specialrule{0em}{2pt}{2pt} 
    
    \tabincell{l}{{\texttt{pendulum}} \\ {\texttt{swingup}}} & \tabincell{c}{10.4 \\ \footnotesize{$\pm$20.6} } & \tabincell{c}{\textbf{33.8} \\ \footnotesize{$\pm$37.0}} & \tabincell{c}{133.5 \\ \footnotesize{$\pm$148.1}} & \tabincell{c}{\textbf{303.6} \\ \footnotesize{$\pm$223.1}} & \tabincell{c}{{306.0} \\ \footnotesize{$\pm$272.9}} & \tabincell{c}{\textbf{471.9} \\ \footnotesize{$\pm$179.7}} & \tabincell{c}{{1.4} \\ \footnotesize{$\pm$1.3}} & \tabincell{c}{\textbf{20.9} \\ \footnotesize{$\pm$23.7}} & \tabincell{c}{38.9 \\ \footnotesize{$\pm$30.5}} & \tabincell{c}{\textbf{45.1} \\ \footnotesize{$\pm$39.4}} & \tabincell{c}{{30.4} \\ \footnotesize{$\pm$26.5}} & \tabincell{c}{\textbf{42.1} \\ \footnotesize{$\pm$23.3}}\\

   \specialrule{0em}{2pt}{2pt}  
   
   \tabincell{l}{\texttt{reacher} \\ \texttt{easy}} & \tabincell{c}{363.4 \\ \footnotesize{$\pm$113.2} } & \tabincell{c}{\textbf{397.3} \\ \footnotesize{$\pm$117.2}} & \tabincell{c}{300.7 \\ \footnotesize{$\pm$127.7}} & \tabincell{c}{\textbf{422.1} \\ \footnotesize{$\pm$136.1}} & \tabincell{c}{380.4 \\ \footnotesize{$\pm$209.7}} & \tabincell{c}{\textbf{437.2} \\ \footnotesize{$\pm$137.9}} & \tabincell{c}{335.9 \\ \footnotesize{$\pm$117.6}} & \tabincell{c}{\textbf{357.1} \\ \footnotesize{$\pm$111.7}} & \tabincell{c}{297.6 \\ \footnotesize{$\pm$168.0}} & \tabincell{c}{\textbf{468.7} \\ \footnotesize{$\pm$130.4}}  & \tabincell{c}{350.3 \\ \footnotesize{$\pm$205.1}} & \tabincell{c}{\textbf{434.3} \\ \footnotesize{$\pm$147.7}} \\

   \specialrule{0em}{2pt}{2pt} 
    
   \tabincell{l}{\texttt{walker} \\ \texttt{walk}} & \tabincell{c}{511.9 \\ \footnotesize{$\pm$63.8} } & \tabincell{c}{\textbf{553.1} \\ \footnotesize{$\pm$47.4}} & \tabincell{c}{591.7 \\ \footnotesize{$\pm$77.0}} & \tabincell{c}{\textbf{630.0} \\ \footnotesize{$\pm$71.9}} & \tabincell{c}{\textbf{644.6} \\ \footnotesize{$\pm$66.0}} & \tabincell{c}{620.2 \\ \footnotesize{$\pm$60.2}} & \tabincell{c}{257.7 \\ \footnotesize{$\pm$103.2}} & \tabincell{c}{\textbf{318.2} \\ \footnotesize{$\pm$70.3}} & \tabincell{c}{303.0 \\ \footnotesize{$\pm$107.7}} & \tabincell{c}{\textbf{327.9} \\ \footnotesize{$\pm$89.7}} & \tabincell{c}{{331.3} \\ \footnotesize{$\pm$80.5}} & \tabincell{c}{\textbf{333.1} \\ \footnotesize{$\pm$53.0}}\\
   
    \bottomrule
    \end{tabular}%

  }
\end{table*}%

\begin{figure*}[t]
  \centering
  \includegraphics[width=17cm]{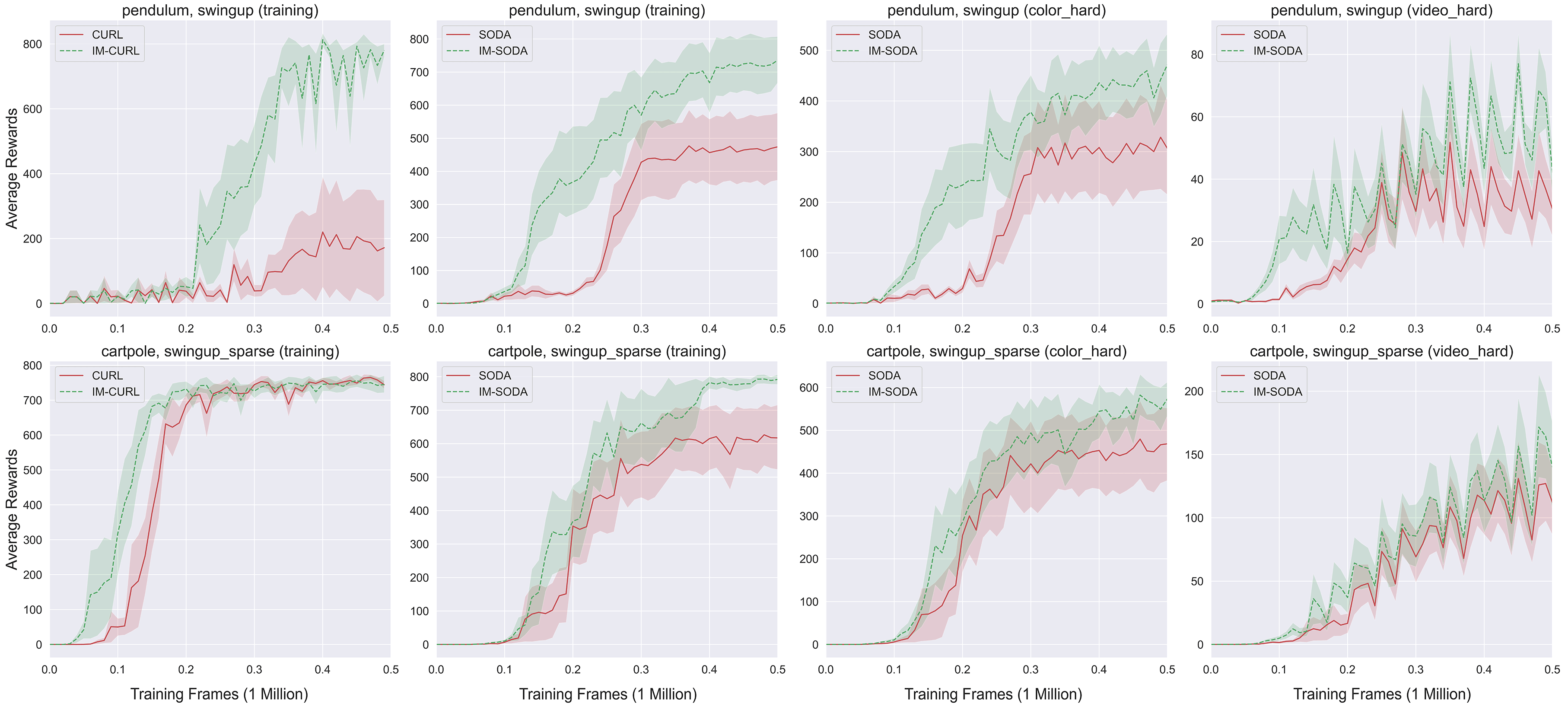}
  \caption{Evaluation with sparse rewards on CURL, IM-CURL, SODA and IM-SODA (8 seeds).}
    \label{fig:sparse}
\end{figure*}

As shown in Table~\ref{generalization}, in the early-term and mid-term, IM-SODA outperforms SODA nearly on all tasks, which proves that its generalization ability in unseen challenging evaluation environments is much better. Since \texttt{video\_hard} is relatively harder, it makes both algorithms perform poorly in some environments. 
Still, the peak of IM-SODA is higher than SODA, which also shows better generalization ability of IM-SODA.

\subsection{Sparse Cases}\label{section:sparse}

Although methods like CURL and SODA achieve comparable performance in major tasks in DMControl, they still perform poorly on tasks with sparse rewards like \texttt{cartpole\_swingup\_sparse} and \texttt{pendulum\_swingup}. As discussed above, simply combining RL with self-supervised learning cannot help solve environments with sparse rewards, but the agent of IM-SSR is encouraged to explore for novel states that have non-trivial rewards potentially.

To verify this hypothesis, we visualize the training process on these two environments with sparse reward signals in Figure~\ref{fig:sparse}. 
Each row in Figure~\ref{fig:sparse} corresponds to an environment, \texttt{pendulum\_swingup} is on the top, while \texttt{cartpole\_swingup\_sparse} is on the bottom. The first column shows the comparison between CURL and IM-CURL on \texttt{training}, while other columns on the right show the comparison between SODA and IM-SODA on \texttt{training}, \texttt{color\_hard} and \texttt{video\_hard} respectively. IM-SSR improves the sample efficiency significantly in almost all cases. Specifically, the result on the left top panel shows that in IM-CURL \texttt{pendulum\_swingup} converges to much higher averaged returns at about 400K steps, while CURL still performs extremely bad at 500K steps.

\begin{figure*}[t]

  \centering
    \includegraphics[width=17cm]{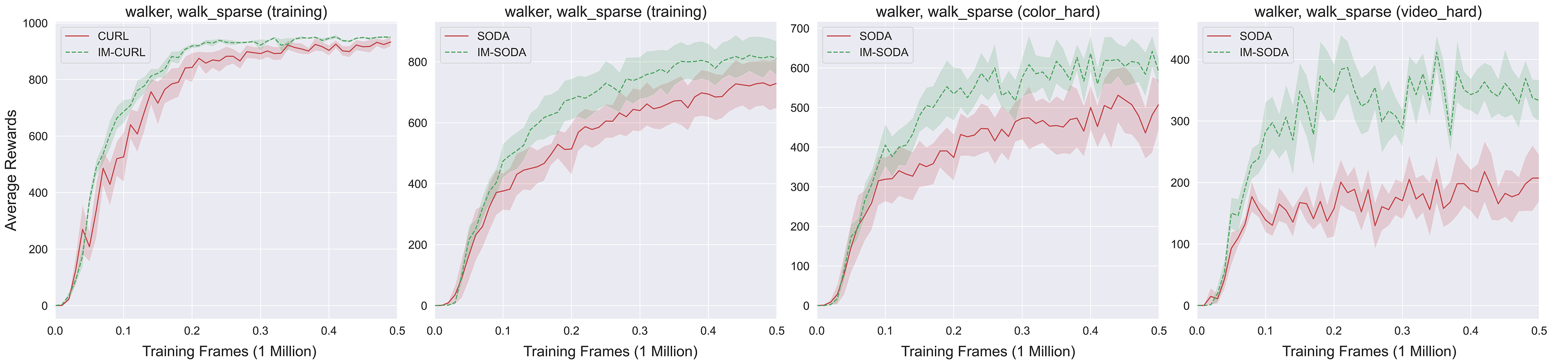}
  \caption{Elaborately designed sparse rewards on CURL, IM-CURL, SODA and IM-SODA in \texttt{walker\_walk\_sparse} (4 seeds).}
    \label{fig:sparse_2}

\end{figure*}

\textbf{What makes the improvement to be marginal or significant?} As we notice in Table \ref{sample}, sometimes IM-SSR and the corresponding baselines achieve similar performance. The decomposition and interpretation of pair-wise contrastive loss in Equation (\ref{finalbound}) can be an explanation, especially the exploration term. When the reward signal is dense, the baseline already learns well without exploring for novel states, and there is not much room for IM-SSR to improve. However, when the reward signal is sparse, the significance of exploration emerges and the intrinsic reward does help. 

To verify this idea, we take \texttt{walker\_walk} as an example and modify the reward to be sparse as either 0 or 1. The new task is called \texttt{walker\_walk\_sparse}. In \texttt{walker\_walk}, IM-SSR and the corresponding baselines have already achieved high performance in early-term. However, as shown in Figure~\ref{fig:sparse_2} with designed sparse settings, IM-CURL and IM-SODA perform much better than CURL and SODA due to the lack of reward supervision.

%% file: sec5_related.tex
\section{Related Work}

\textbf{Self-Supervised Learning for Visual Representation.} Self-supervised learning has attracted interests of researchers in Computer Vision~\cite{he2020momentum, chen2020simple, grill2020bootstrap, oord2018representation, chen2020exploring, zbontar2021barlow}. \rebuttal{MoCo (Momentum Contrastive)~\cite{he2020momentum}} builds a dynamic queue with the help of the momentum encoder. It trains the encoder by matching queries and keys using InfoNCE loss. SimCLR~\cite{chen2020simple} introduces a learnable nonlinear transformation between the representation and the contrastive loss. With the projection layers, we can remove the momentum encoder and the memory bank. BYOL~\cite{grill2020bootstrap} trains an online network to predict the target network representation of the same image with the different views. Amazingly, they remove negative samples. SimSiam~\cite{chen2020exploring} explores what makes good representation in BYOL by modifying the predictor and removing the momentum encoder. 

\textbf{Auxiliary Task in Reinforcement Learning.}
Directly using images as the input in deep reinforcement learning always causes sample inefficiency, researchers propose to jointly train the policy and a self-supervised loss to solve the inefficiency problem to improve performances~\cite{shelhamer2016loss, hansen2020generalization, srinivas2020curl, yarats2019improving,  hansen2020self, hansen2021stabilizing}.
Inspired by \cite{he2020momentum, chen2020simple}, CURL~\cite{srinivas2020curl} uses contrastive learning as the auxiliary tasks. They just \rebuttal{treat the different randomly cropped views} of the same image as the positive pair. SODA~\cite{hansen2020generalization} introduces a new overlay augmentation that they use the augmentation and \rebuttal{randomly crop the} same image as the positive pair, which shows State-of-The-Art performance in generalization \cite{tassa2018deepmind}. \rebuttal{In} \cite{grill2020bootstrap}, they use a online network to predict the target network and do not need negative pairs. Besides, they do not use the same batch data to optimize the RL loss and the auxiliary task loss. 
PAD~\cite{hansen2020self} \rebuttal{uses} inverse dynamics prediction error to optimize the encoder and introduces policy adaptation tricks to improve generalization performance. Besides, SVEA~\cite{hansen2021stabilizing} stables  the learning of Q-value to improve generalization. However, \rebuttal{the policy training part and SSL part} are separated, which prevents the policy from referencing to representation learning to make decision. 

\textbf{Exploration.}
Exploration is an important topic in reinforcement learning. A natural and effective way is \rebuttal{count-based} exploration bonuses~\cite{strehl2008analysis}. When \rebuttal{the} space is becoming larger and larger, simple \rebuttal{counting} fails, and lots of work have studied how to generalize count exploration to large state spaces~\cite{burda2018exploration, bellemare2016unifying, fu2017ex2}. 
Another class of exploration is based on the errors in predicting dynamics~\cite{burda2018exploration, schmidhuber1991possibility, burda2018large}. For example, \rebuttal{Random Network Distillation~\cite{burda2018exploration}} treats the error of predicting features of the next observations given by a fixed random model as the bonus. When policy gives an action to find a non-seen state, it will be awarded. However, exploration based on predicted errors needs additional network architectures, while our method only borrows the representation learning error as the intrinsic reward.

%% file: sec6_conclusion.tex
\section{Conclusions and Future Work}


In summary, we present a simple yet effective idea to introduce the information in self-supervised learning into policy learning as intrinsically motivated self-supervised learning in reinforcement learning (IM-SSR). Theoretical analysis is made to interpret decomposition of the self-supervised loss, as exploration for novel states and robustness from nuisance elimination. The implementation of the IM-SSR is quite simple, without any extra modification on architectures nor costs in computation, yet the improvement is significant. Both IM-CURL and IM-SODA achieve faster convergence than CURL and SODA. Besides, IM-SODA generalizes much better than SODA in unseen real-world like environments. We also emphasize the IM-SSR performs much better than the underlying baselines in sparse reward environments.
Therefore, we are offering an approach to further improve sample efficiency and generalization of suitable SSL-RL baselines, with nearly no additional cost.

Albeit it is promising to introduce interaction between SSL and RL, intrinsic reward is not the only way to implement this idea; thereby, many other approaches remain to be developed. Also, an adaptive schedule can be designed for the intrinsic parameter $\beta_t$ to achieve a more robust and tuning-free algorithm. 
In the future, we desire to further extend IM-SSR to more robotics tasks in more complex environments.

%% file: appendix.tex


\appendices

\begin{figure*}[!thbp]
\centering
\includegraphics[width=17cm]{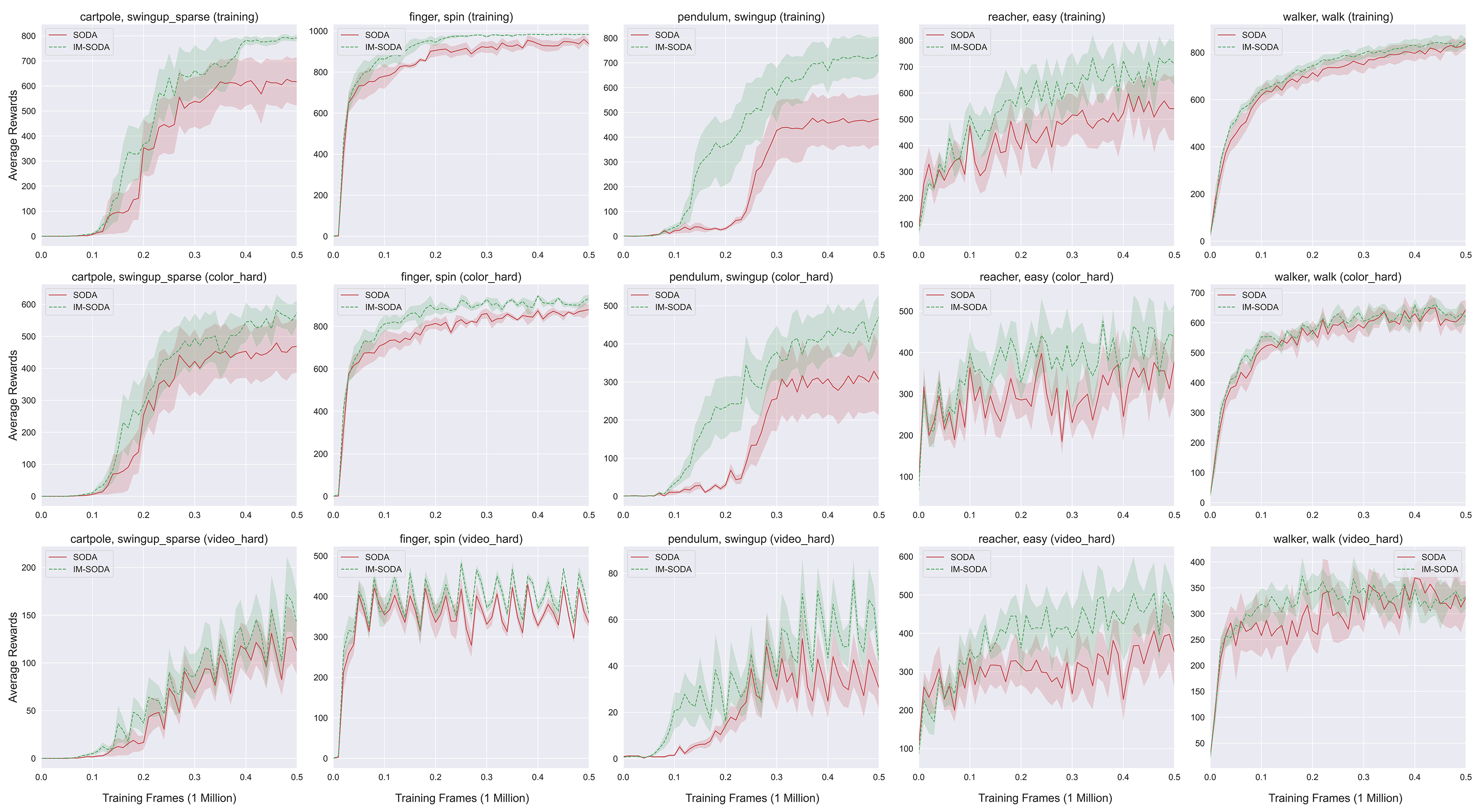}
\caption{Comparison between SODA and IM-SODA (\rebuttal{8 seeds in \texttt{cartpole} and \texttt{pendulum}})}
\end{figure*} 

 

\section{Performance Curves}\label{ref-performance}

To show the effectiveness of IM-SSR, we visualize the performance curves in DMControl environments. To further verify the generalization ability of IM-SSR from simulation to real world, evaluation on complex unseen environments is conducted based on DMControl-GB. As we can see, almost in all environments, IM-SSR outperforms the baseline methods. We mainly follow the experimental setting in CURL and SODA, which trains the agent in unmodified \texttt{training} environments and test in various environments: unmodified \texttt{training}, \texttt{color\_hard} and \texttt{video\_hard}. The x-axis denotes training frames (1 million).

\begin{figure*}[htbp]
\centering
\includegraphics[width=17.5cm]{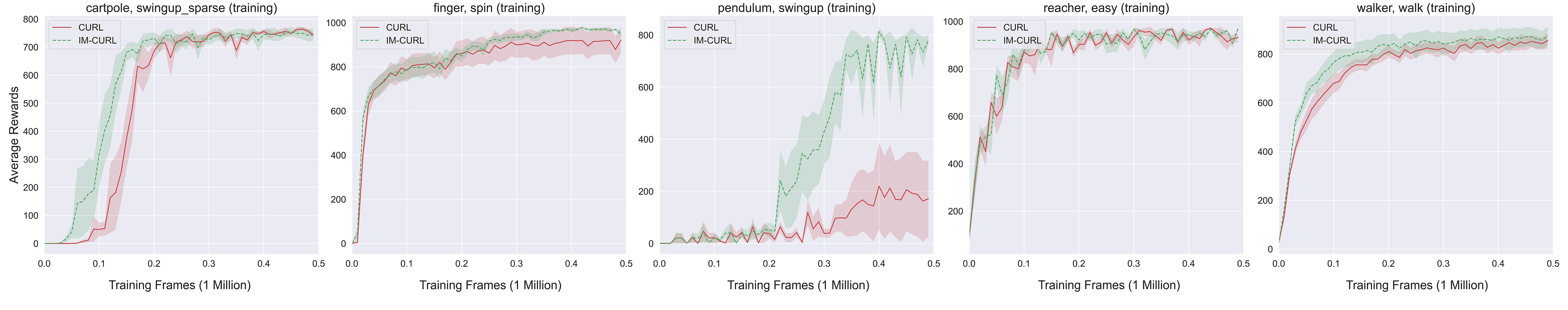} 
\caption{Comparison between CURL and IM-CURL}
\end{figure*} 

\begin{figure*}[htbp]
\centering
\includegraphics[width=17.5cm]{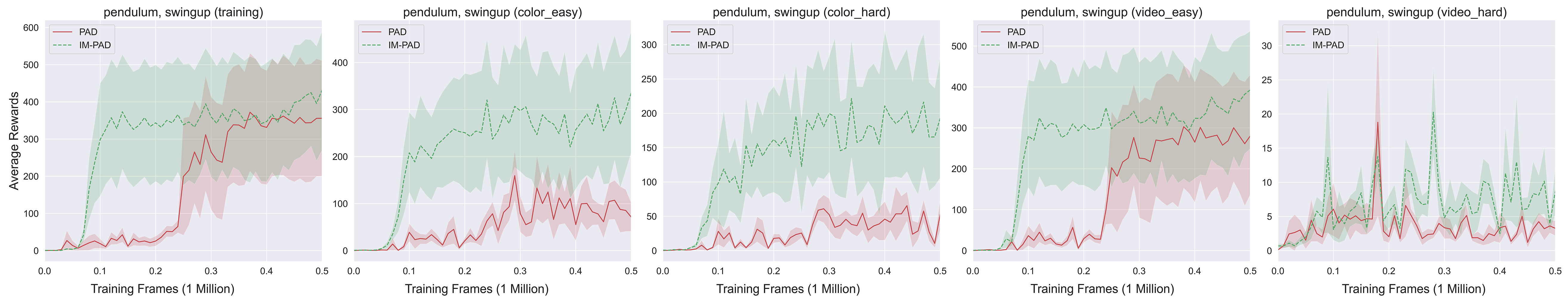}
\caption{Comparison between PAD and IM-PAD}
\label{fig:pad}
\end{figure*}

\rebuttal{Additional results related to PAD and SVEA are in Figure~\ref{fig:pad} and Figure~\ref{fig:svea}, where $\beta_0=0.05$. We focus on the performance on \texttt{pendulum\_swingup}.}

The performance curves clearly demonstrate the priority of IM-SSR on sample efficiency over baselines. The green curve which represents IM-SSR is higher than the red baseline almost every time. Especially in tasks with sparse rewards, like the first and the third columns, IM-SSR obtains non-trivial rewards much earlier and surpasses the underlying baseline significantly. Also in generalization evaluation, IM-SODA shows better generalization ability in \texttt{color\_hard} and \texttt{video\_hard}.

\section{Basic SSL-RL Architectures}\label{ssl-formulation}
Our method can be easily deployed on any SSL-RL framework. In this paper, we use CURL~\cite{srinivas2020curl}, SODA~\cite{hansen2020generalization} \rebuttal{and PAD~\cite{hansen2020self}} as the basic SSL architectures in the IM-SSR framework, which are built based on SAC~\cite{haarnoja2018soft}.

\textbf{CURL.}
Vision-based RL is disturbed by sample-inefficiency~\cite{yarats2019improving}. Inspired by the success of self supervised learning in computer vision~\cite{he2020momentum,chen2020simple}, CURL~\cite{srinivas2020curl} proposes to training an extra contrastive objective as an auxiliary task to make encoder to learn high level representation faster and better. More specifically, they maximize the agreement between augmented versions of the same observation by InfoNCE loss as Equation~(\ref{infonce}):
\begin{equation}
\label{infonce}
\mathcal{L}_{\text{CURL}}=-\log \frac{\exp \left(q^{T} W k_{+}\right)}{\exp \left(q^{T} W k_{+}\right)+\sum_{i=0}^{K-1} \exp \left(q^{T} W k_{i}\right)},
\end{equation}
where $q$ is the encoder query, $W$ is parameters to be optimized and $k_{+}$ denotes the key that matches $q$~\cite{srinivas2020curl}. Intuitively, the value of the loss is low when $q$ is similar to its positive key $k_{+}$ and dissimilar to other negative keys~(denoted as $k_{i}$). In CURL implementation, we simply treat different augmented views of the same state as the positive pair.

\textbf{SODA.}
SODA~\cite{hansen2020generalization} maximizes the mutual information between latent representations of augmented and non-augmented data by employing BYOL-like~\cite{grill2020bootstrap} architecture, which does not need negative samples at all. The RL policy and the self-supervised auxiliary task share a common encoder $f_\theta$. $o^{\prime}$ is an augmented observation of $o$. They use the encoder $f_\theta$ and an projection network $g_\theta$ to extract $z^{\prime}=g_\theta(f_\theta(o^{\prime}))$, and use the target encoder $f_\psi$ and target projection network $g_\psi$ to extract $z^{\star}=g_\psi(f_\psi(o^{\star}))$, where $\psi$ is an exponential moving average~(EMA) of $\theta$. The objective of SODA is to predict $z^{\star}$ from $z^{\prime}$ by $h_{\theta}$, which is formulated as a consistency loss:
\begin{equation}
\label{sodaloss}
\mathcal{L}_{\text{SODA}}\left(\hat{z}, z^{\star} ; \theta\right)=\mathbb{E}_{t \sim \mathcal{T}}\left[\left\|\hat{z}_{\circ}-z_{\circ}^{\star}\right\|_{2}^{2}\right],
\end{equation}
where $\hat{z}\triangleq h_\theta(z^{\prime})$, $\hat{z_o}\triangleq \hat{z}/\left\|\hat{z}\right\|_2$ and $z_o^\star\triangleq z^\star/\left\|z^\star\right\|_2$.

\rebuttal{\textbf{PAD.} PAD~\cite{hansen2020self} explores to use self-supervision in new environments for continue training the policy for generalization. They employ the inverse dynamics model as the SSL part. Specifically, at each step, a transition sequence ($s_t$,$a_t$,$s_{t+1}$) is observed and an inverse model takes $s_t$ and $s_{t+1}$ to predict $a_t$, which can adapt the policy to the new environments without any reward signal. Formally, the inverse dynamics objective for continuous actions can be written as:
\begin{equation}
    \mathcal{L}_{\text{PAD}} = MSE(a_t, f(s_t, s_{t+1})) 
\end{equation}
where MSE means mean squared error and $f$ means the inverse prediction model.}

\textbf{SVEA.} SVEA~\cite{hansen2021stabilizing} improves the generalization of RL algorithms by stabling the learning of the Q-values. To be more specific, SVEA minimizes a nonnegative objective $\mathcal{L}_{\text{SVEA}}$, where

\begin{equation}
\begin{aligned}
\mathcal{L}_{\text{SVEA}}=\mathbb{E}& { [\alpha\left\|Q\left(f\left(s_{t}\right), a_{t}\right)-q_{t}^{\text {target }}\right\|} \\
&+\beta\left\|Q\left(f\left(s_{t}^{a u g}\right), a_{t}\right)-q_{t}^{\text {target }}\right\| ]
\end{aligned}
\end{equation}

$q_t^{target}$ is the target q-value in Bellman equation. $\alpha$ and $\beta$ is the constant coefficients and $s^{aug}_t$ means we use the augmented data to compute the Q-values.

\textbf{Data Augmentations.} In order to smoothly apply SSL to RL tasks, we need to augment the same observation. CURL only augments data by random crop, while SODA proposes a novel and stronger augmentation method, random overlay, which linearly interpolates between an observation and another image. In CURL and PAD, the inputs of different encoders are different crops of the same observation; in SODA, both inputs are the same crop, while the momentum encoder's input is additionally processed by the random overlay augmentation method. More details can be found in \cite{hansen2020generalization, hansen2020self, srinivas2020curl}. \rebuttal{Noting that our IM-SSR uses the same augmentation method as baselines.}

\rebuttal{\textbf{Update Details.} Besides, there are differences in the update of auxiliary tasks: Some, like CURL, update the auxiliary task with same sampled transitions as in RL update, while others do not, like SODA. We find it interesting that using different transitions improves the performance, which remains to be further studied. Also, the update frequency of RL and SSL can be different. We mention these differences in the pseudo-code and strictly follow the same way as the corresponding baseline does.}

\section{Implementation Details}\label{ref-implementation}
{\subsection{Baseline Details}}
We show the implementation details for CURL, SODA and \rebuttal{PAD} in this subsection. Specifically, we present in detail about the hyperparameters for both algorithms in Table~\ref{tb:details} and the choice of the $\beta$ for intrinsic reward in Table~\ref{tb:gamma}. Noting that the basic SSL-RLs are mainly based on the official released implementation\footnote{\url{ https://github.com/MishaLaskin/curl}} \footnote{\url{ https://github.com/nicklashansen/dmcontrol-generalization-benchmark}}.

We utilize a simple decaying schedule for $\beta$, by multiplying a rate every 100k steps, like $\beta \leftarrow \beta * 0.8$. All experiments follow this rule, except for \texttt{cartpole\_swingup\_sparse}, which uses a fixed parameter $\beta = \beta_0$ without any complicated designed schedule. Such a hyper-parameter, which is fixed or naively scheduled, can still obtain solid performance, thus proving the stability of IM-SSR. In the future, adaptive schedule can be used to adjust $\beta$. Besides, similar to RND \cite{burda2018exploration}, we use the predicted observation error of the next state.

\begin{figure}[htbp]
\centering
\subfigure[raw input]{
\begin{minipage}[t]{0.25\linewidth}
\centering
\includegraphics[width=1in]{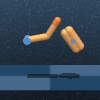}
\end{minipage}%
}%
\subfigure[natural image]{
\begin{minipage}[t]{0.25\linewidth}
\centering
\includegraphics[width=1in]{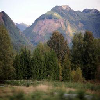}
\end{minipage}%
}%
\subfigure[overlay]{
\begin{minipage}[t]{0.25\linewidth}
\centering
\includegraphics[width=1in]{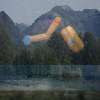}
\end{minipage}
}%

\centering
\caption{Data Augmentation: Random Overlay}
\label{random_overlay}
\end{figure}

\rebuttal{For data augmentation, both CURL and PAD only use random cropping while SODA proposes a new augmentation method for generalization: Random Overlay~\cite{hansen2020generalization}, which linearly interpolates between an observation and another natural image as Figure~\ref{random_overlay} shows. We can formally write it as:
\begin{equation}
    t_{overlay} = (1-\alpha) * o + \alpha * \epsilon,
\end{equation}
where $o$ is the observation, $\epsilon$ is the natural image and $\alpha$ is the interpolation coefficient. In SODA~\cite{hansen2020generalization}, $\alpha$ is set as $0.5$. We emphasize that we use exactly the same augmentations as the baseline methods.}

{\subsection{Hyper-parameters}}

Hyper-parameters are summarized in Table~\ref{tb:details} and Table~\ref{tb:gamma}.
\begin{table*}[t]
\centering
\caption{The Hyper-parameters used in our experiments.}
\begin{tabular}{ll}
\toprule
Hyper-parameter & Value \\
\specialrule{0.05em}{2pt}{2pt} 

Frame rendering & $3 \times 100 \times 100$ \\
Frame after crop & $3 \times 84 \times 84$ \\
Stacked frames & 3 \\
Number of conv. layers & 11 (SODA\rebuttal{, PAD}, SVEA) \\
& 4 (CURL)\\
Number of filters in conv. & 32 \\
Action repeat & 2 (\texttt{finger\_spin}) \\
& 8 (\texttt{cartpole\_swingup\_sparse}, \texttt{pendulum\_swingup}) \\
& 4 (otherwise) \\
Discount factor $\gamma$ & $0.99$ \\
Episode time steps & 1,000 \\
Learning algorithm & Soft Actor-Critic \\
Number of training steps & 500,000 \\
Replay buffer size & 500,000 (SODA\rebuttal{, PAD}, SVEA) \\
& 100,000 (CURL) \\
Optimizer (RL/aux.) & Adam $\left(\beta_{1}=0.9, \beta_{2}=0.999\right)$ \\
Optimizer $(\alpha)$ & Adam $\left(\beta_{1}=0.5, \beta_{2}=0.999\right)$ \\
Learning rate (RL) & 1e-3 \\
Learning rate $(\alpha)$ & 1e-4 \\
Learning rate $($ SODA $)$ & 3e-4 \\
Batch size in RL & 128 \\
Batch size in SSL & 128 (CURL)\\
 & 256 (SODA\rebuttal{, PAD}, SVEA) \\
Actor update freq. & 2 \\
Critic update freq. & 1 \\
Auxiliary update freq. & 1 (CURL\rebuttal{, PAD}) \\
& 2 (SODA) \\
Momentum coef. $\tau$ (SODA) & $0.005$\\
\bottomrule
\end{tabular}
\label{tb:details}
\end{table*}

\begin{table*}[!t]
\centering
\caption{The initial $\beta_0$ of Intrinsic Reward in different environments.}
\begin{tabular}{lllllll}
\toprule
 & \texttt{cartpole} & \texttt{finger} & \texttt{pendulum} & \texttt{reacher} & \texttt{walker} & \rebuttal{\texttt{walker}}\\
 & \texttt{swingup\_sparse} & \texttt{spin} & \texttt{swingup} & \texttt{easy} &  \texttt{walk} &  \rebuttal{\texttt{walk\_sparse}}\\
\specialrule{0.05em}{2pt}{2pt} 

IM-CURL & 0.01 & 0.005 & 0.05 & 0.005 & 0.005 & 0.1 \\
IM-SODA & 0.01 & 0.005 & 0.1 & 0.001 & 0.05 & 0.1 \\
\bottomrule
\end{tabular}
\label{tb:gamma}
\end{table*}

\subsection{Sparse Reward Settings}

\rebuttal{In Section~\ref{section:sparse}, we modify the reward signal in \texttt{walker\_walk} to compare the improvements between dense rewards and sparse rewards. The original reward signal designed in DMControl Suite is a combination of terms related to upright torso, torso height and horizontal velocity~\cite{tassa2018deepmind}. The returned reward is scaled from 0 to 1, which is computed by the following codes.}

\begin{minted}[frame=lines, breakanywhere, breaklines]{Python}
def get_reward(self, physics):
  """Returns a reward to the agent."""
  standing = rewards.tolerance(
    physics.torso_height(),
    bounds=(_STAND_HEIGHT, float('inf')),
    margin=_STAND_HEIGHT/2)
  upright = (1+physics.torso_upright())/2
  stand_reward = (3*standing+upright)/4
  if self._move_speed == 0:
    return stand_reward
  else:
    move_reward = rewards.tolerance(
      physics.horizontal_velocity(),
      bounds=(self._move_speed, float('inf')), 
      margin=self._move_speed/2,
      value_at_margin=0.5,
      sigmoid='linear')
    return stand_reward*(5*move_reward+1)/6
\end{minted}

\rebuttal{To maintain the desired combination of rewards in walking, we keep the same setting of the dense reward, and only change it from the continuous version in $[0, 1]$ to a discrete version in $\{0, 1\}$. The sparse reward is defined as follows.
\begin{equation}
\text{sparse\_reward} =
\begin{cases}
1.0 & \text{dense\_reward > 0.5}\\
0.0 & \text{otherwise}
\end{cases}
\end{equation}
}

{\section{Further Extensions}\label{app:imsvea}}

Various architectures of SSL-RL can also be explored, such as SVEA~\cite{hansen2021svea} whose self-supervised learning part is merged into the Q-learning procedure. Hence, we cannot directly utilize the SSL loss as an intrinsic reward, since no auxiliary loss is available.

However, the philosophy of IM-SSR to utilize the paramount information in SSL can still be realized by information maintained in encoders. We simply add an MSE loss as the metric to evaluate the distance between encoded augmentations and encoded original observations, and then use the pair-wise MSE as the intrinsic reward. 
The additional computation is still very little.

The detailed implementation is as follows. $\hat{z}_{i}$ denotes the encoded variable of the original observation by encoder in Q function, while $z_{i}^{\star}$ denotes the encoded variable of the original observation by the encoder in target function. By this way, we can calculate $l_i = \left\|\hat{z}_{i}-z_{i}^{\star}\right\|_{2}^{2}$ and normalize them within a batch to design the intrinsic reward.

\begin{figure*}[t]
\centering
\includegraphics[width=17cm]{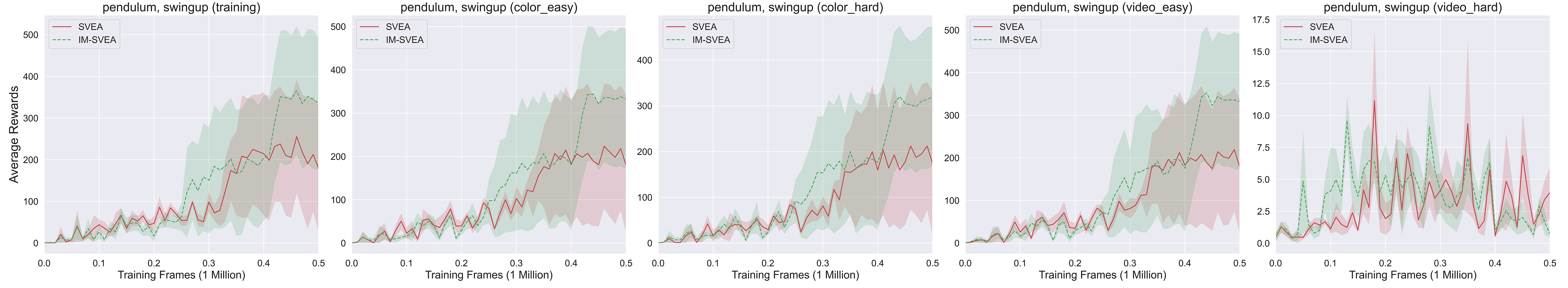} 
\caption{Comparison between SVEA and IM-SVEA}
\label{fig:svea}
\end{figure*}

The results show that IM-SVEA achieves better performance than SVEA in Figure~\ref{fig:svea}. The improvements in \texttt{video\_hard} can be ignored, which may be attributed to the baseline SVEA who fails.

This section motivates us to explore various approaches to design the intrinsic reward, when the formulation of SSL loss is not desired or even there is no available auxiliary loss. More attempts on how to efficiently utilize the information in the SSL part are needed in the future.

\section{Generalization Benchmarks}\label{transfer-description}

\subsection{Motivations on Generalization Tasks}
\rebuttal{We include two types of experiments related to generalization, one is more likely to sim-to-real, and another is more likely to transferring. In Section~\ref{section-gen}, we train agents based on unmodified environments and tested on unseen challenging environments which is designed for simulation to reality. We also implemented experiments that are trained on one scene and tested on another unseen scene with a dynamical changing camera pose, which is designed for transferring in real world.}

\rebuttal{First, we would like to explain for generalization tasks in Section~\ref{section-gen} based on DMControl-GB. Researchers mainly focus on the original DMControl baseline in previous works, which is still away from real life. If we have a task in the real world, sometimes we do not train directly in the real world; we can model the object and simulate the task in a mujoco-like world instead. The physical objects are already included in the simulated environment, and the main differences will be nuisances we mentioned in the paper, like color, background, and others. Therefore, it is reasonable to treat the benchmark as a useful generalization benchmark, which helps transfer from simulation to real life.}

\begin{figure}[t]
  \centering
  \subfigure{\includegraphics[scale=0.5]{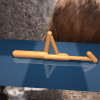}} \hspace{.05in}
  \subfigure{\includegraphics[scale=0.5]{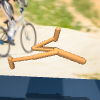}}\hspace{.05in}
  \subfigure{\includegraphics[scale=0.5]{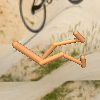}}\hspace{.05in}
    \subfigure{\includegraphics[scale=0.5]{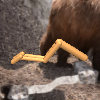}}
  \caption{Various Scenes in Distracting Control Suite Environments \cite{stone2021distracting}}
  \label{fig:envs-dis}
\end{figure}

In this work, to the best of our ability, we try to simulate transferring from one scene to another. The implementation of transferring tasks is based on DMControl Generalization Benchmark~\cite{hansen2020generalization} and Distracting Control Suite~\cite{stone2021distracting}. Additional challenging experiments are implemented, where the agent is trained based on natural videos instead of simulated mujoco environments and tested in another unseen scene. Apart from the changing of scenes, camera poses are also dynamic in training and testing as shown in Figure~\ref{fig:envs-dis}.

\subsection{Performance in Transferring Tasks}

\begin{figure}[thbp]
  \centering
  \subfigure{\includegraphics[width=4cm]{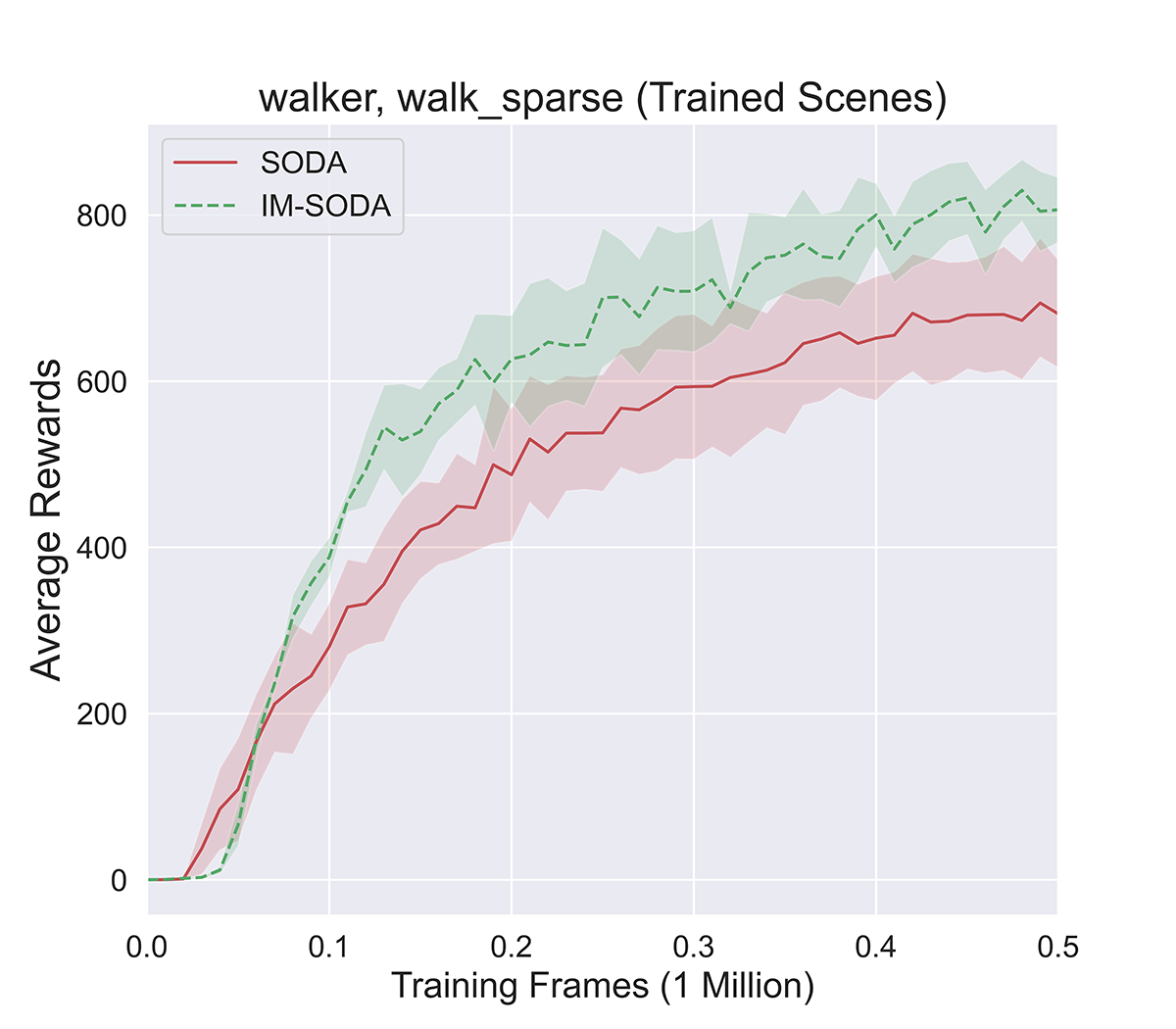}} 
  \subfigure{\includegraphics[width=4cm]{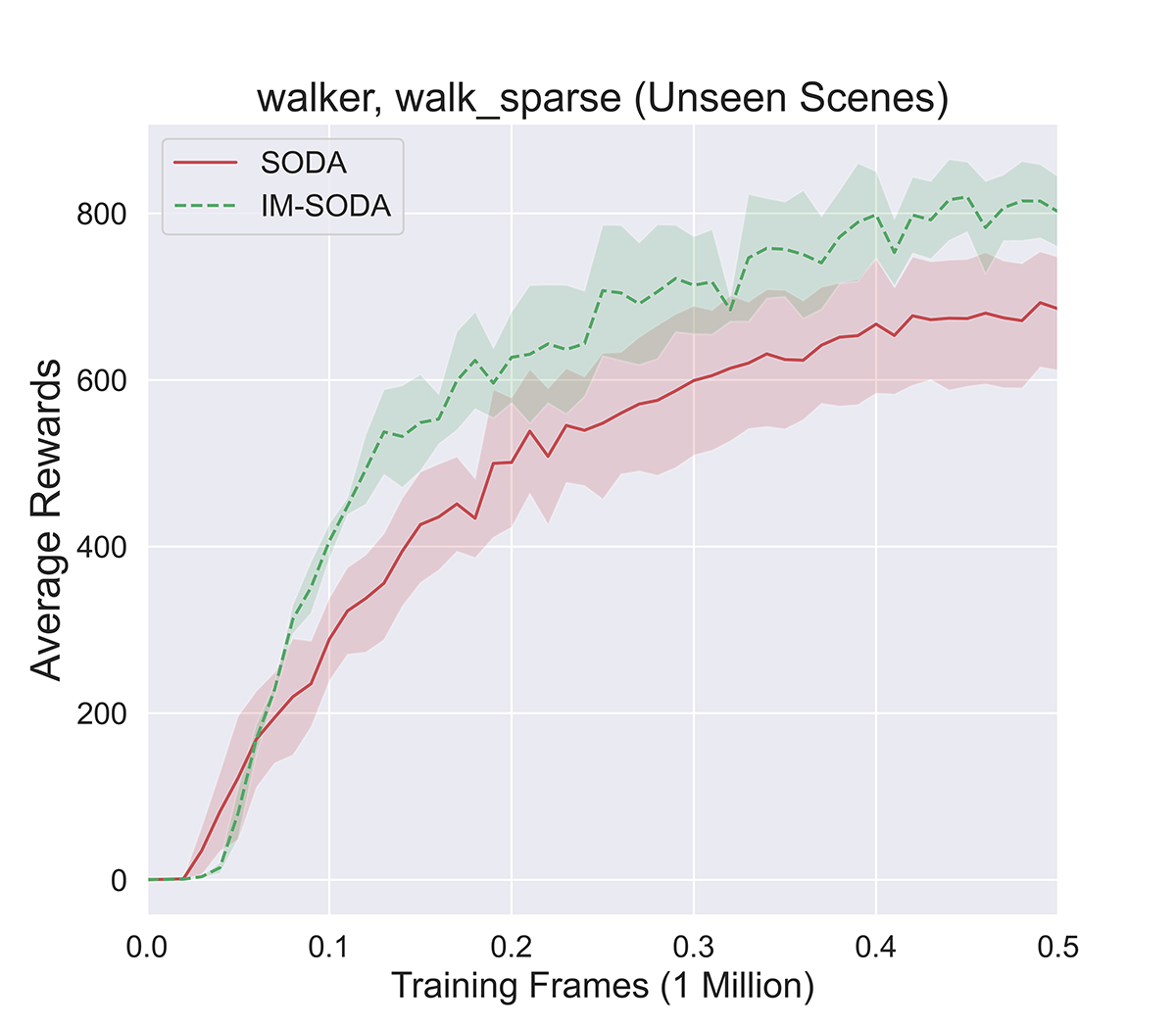}}
\caption{Comparison between SODA and IM-SODA in Transferring Tasks (\rebuttal{4 seeds})}
\label{exp:transfer}
\end{figure} 

\rebuttal{As is shown in Figure~\ref{exp:transfer}, we conduct experiments on \texttt{walker\_walk\_sparse} to compare between the trained scene and another unseen test scene. The performance show that IM-SODA performs better than SODA as desired.}

\section{Sensitivity Analysis on $\beta_0$}
\begin{figure}[thbp]
  \centering
  \subfigure[Training Curve]{\includegraphics[width=4cm]{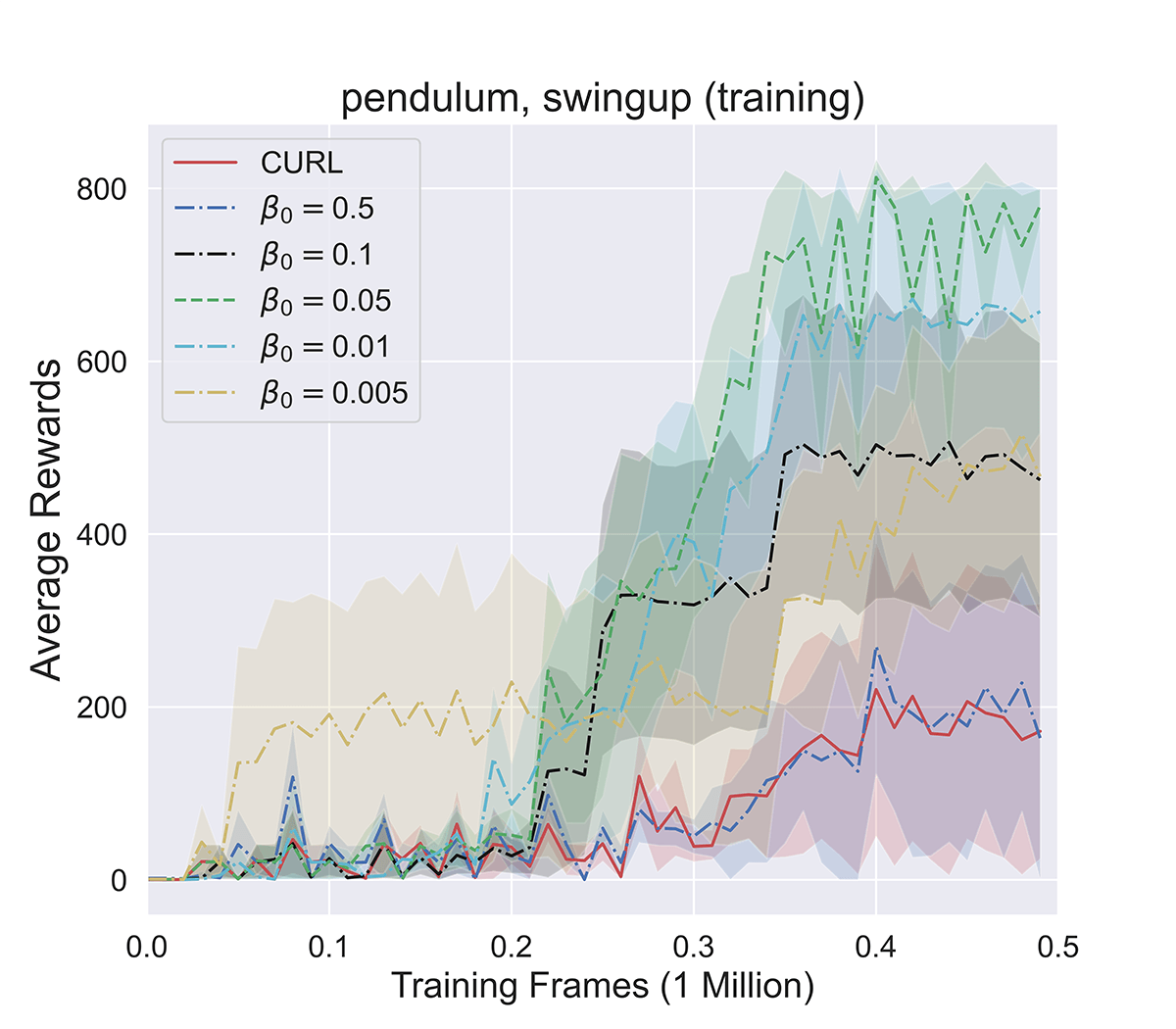}\label{fig:beta1}} 
  \subfigure[Final Performance]{\includegraphics[width=4cm]{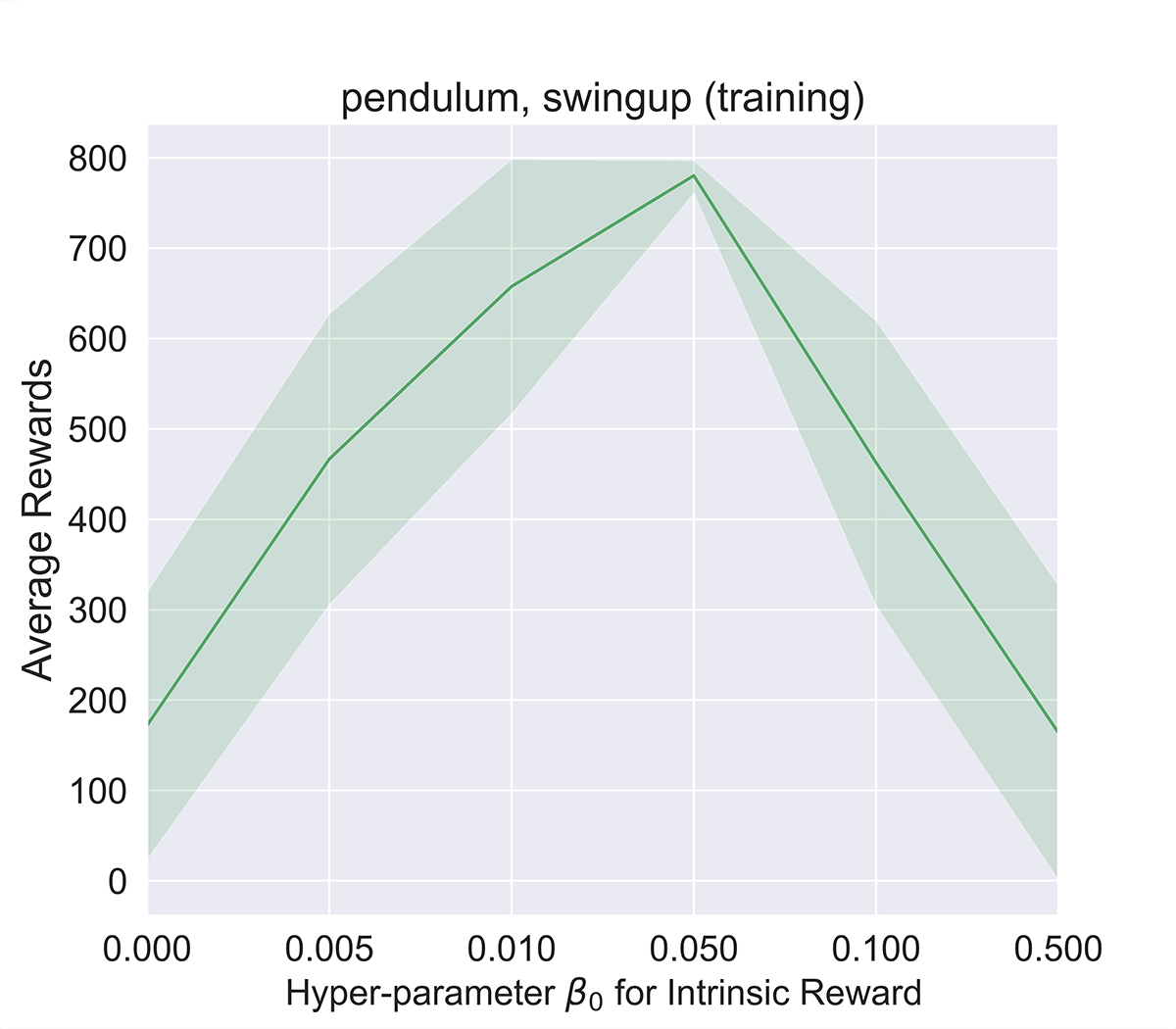}\label{fig:beta2}}
\caption{CURL and IM-CURL with different hyper-parameter $\beta_0$}
\end{figure}

Here, we fix the schedule of $\beta$ and explore the utility of various $\beta_0$ in IM-SSR. Figure~\ref{fig:beta1} shows that different $\beta_0$ will lead to different performance, where a suitable $\beta_0=0.05$ leads to the best performance. As in shown in Figure~\ref{fig:beta2}, with the increasing of $\beta_0$, the final performance will increase first and then decrease. \rebuttal{The magnitude of $\beta_0$ matters for a proper modification on the original reward.} If the $\beta_0$ is too big, the intrinsic reward will be the most important target of the agent, \rebuttal{and the extrinsic reward will be neglected,} which results in low performance. If the $\beta_0$ is too small, the performance is closer to CURL, which is a special case for $\beta_0=0$.

